\pdfoutput=1

\documentclass[11pt]{article}

\usepackage{acl}

\usepackage{times}
\usepackage{latexsym}
\usepackage{float}
\usepackage[T1]{fontenc}

\usepackage[utf8]{inputenc}

\usepackage{microtype}

\usepackage{graphicx}
\usepackage{booktabs}

\title{Controlling for Stereotypes in Multimodal Language Model Evaluation}

\author{Manuj Malik\textsuperscript{\normalfont1}\hspace{6mm}Richard Johansson\textsuperscript{\normalfont2} \\
  \textsuperscript{1}International Institute of Information Technology, Bangalore, India, \\
  \textsuperscript{2}University of Gothenburg and Chalmers University of Technology, Gothenburg, Sweden \\
  \texttt{manuj.malik@iiitb.org, richard.johansson@gu.se} \\}
\begin{document}
\maketitle
\begin{abstract}
We propose a methodology and design two benchmark sets for measuring
to what extent language-and-vision language models use the visual
signal in the presence or absence of stereotypes.
The first benchmark is designed to test for stereotypical colors of common objects, while the second benchmark considers gender stereotypes.
The key idea is to compare predictions when the image conforms to the
stereotype to predictions when it does not.

Our results show that there is significant variation among multimodal
models: the recent Transformer-based FLAVA seems to be more sensitive
to the choice of image and less affected by stereotypes than older
CNN-based models such as VisualBERT and LXMERT. This effect is more
discernible in this type of controlled setting than in traditional
evaluations where we do not know whether the model relied on the
stereotype or the visual signal.
\end{abstract}

\section{Introduction}

The center of gravity of NLP research has shifted to the development
of language models (LMs) for representation and generation of text,
and most recent high-impact research contributions describe new
LMs.
For some tasks, a model needs to take into account not only a text but
also some non-textual information, and a wide range of multimodal LMs
have been developed that allow the representation of a text jointly
with some external modality. Most of this work focuses
on \emph{visual} tasks where NLP models need to be integrated with
computer vision models; examples of tasks in this area include visual
question answering and caption generation.
A range of combined language-and-vision LMs have been developed using
different approaches for integrating representations of text and of
images or videos.

But can we be sure that a multimodal model actually uses the provided visual information instead of just relying on statistical tendencies in the text corpus?
With the development of multimodal LMs, some recent work has
investigated what information is stored in the representations of the
multiple modalities and how the multiple representations interact.
%
For instance, \newcite{frank2021vision} carried out a set of controlled tests
to tease apart the effects of the textual and visual modalities.

\begin{figure}[t]
    \centering
    \includegraphics[width=0.98\linewidth]{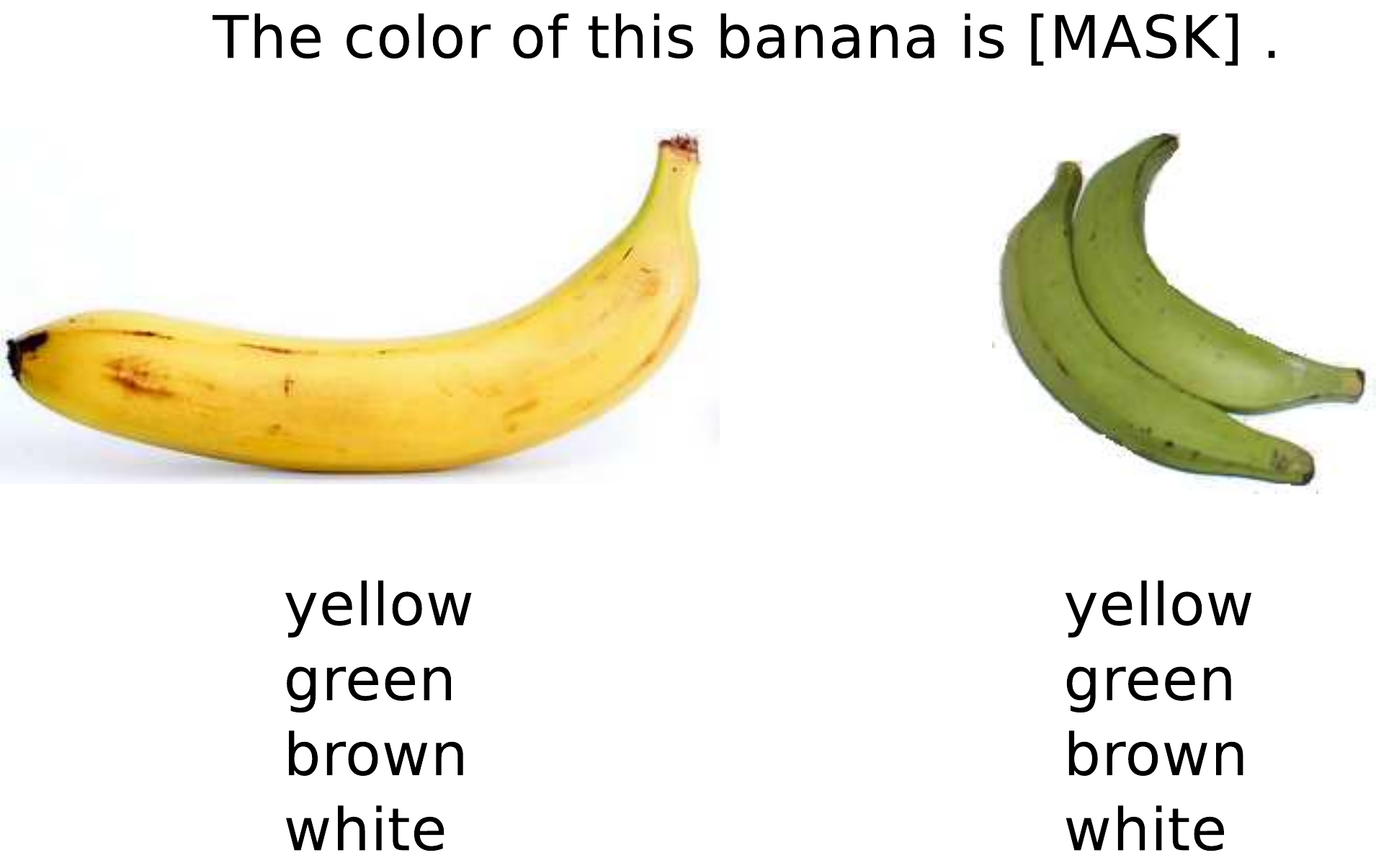}
    \caption{An example of a controlled test of a masked language
    model for a color stereotype. We compute the output from the MLM
    head when providing an image of an object with a stereotypical
    color (a yellow banana) and compare it to the output when the object has an unusual color (green). If the MLM is strongly affected by a stereotype bias, the predictions change little.}
    \label{fig:overview}
\end{figure}

It has been widely noted that representations of language are
affected by several kinds of \emph{stereotypes}, which we loosely
define as any type of phenomenon that has a highly skewed prior
probability distribution.
In these cases, the skewed distribution may cause a model to
simply go with the default choice and
ignore contextual information that would suggest an unusual analysis.
Most of the discussion in the field has been about stereotypes
relating to various demographic attributes \cite{bolukbasi2016}, but
in this work, we use the term ``stereotype'' in the more general sense
mentioned above.
This issue is likely to affect multimodal LMs as well, although we are
aware of no previous work that investigates this phenomenon
systematically; for instance, if some object is often associated with
some visual property (e.g. a color or shape), this property may be
predicted by the model even in cases where it is not present.
This effect may also have methodological implications in benchmarks
for the evaluation of LMs:
if a model predicted the correct answer, did it do so because of the stereotype or because it actually used the available
visual information?

In this work, we propose a methodology and develop two benchmark sets
for stress-testing multimodal LMs to determine to what 
extent they are affected by problems related to stereotypes.
The key idea is to look at predictions of a language/vision LM with
different visual inputs and compare the behavior of the LM in the
presence or absence of stereotypes.
%
For cases when a stereotype is present, we compare model outputs when
the image \emph{does} correspond to the stereotype to when
it \emph{does not}.

The rest of the paper is organized as
follows. Section~\ref{sec:benchmarks} discusses the design of the
benchmark sets and how we use them to investigate multimodal LMs for
stereotypes. Details about the multimodal LMs we have used
are covered in Section~\ref{sec:models}, and
Section~\ref{sec:method} describes how they are applied for the
benchmarks, while 
Section~\ref{sec:results} presents the figures achieved on the
benchmarks and discusses their implications.
In Section~\ref{sec:relwork}, we discuss related research.
Finally, Section~\ref{sec:conclusions} summarizes the main points and
discusses limitations and possible extensions.

\section{Design of Benchmark Datasets}
\label{sec:benchmarks}

We have collected two datasets consisting of textual templates and
corresponding images. These datasets were selected because in these
cases it was relatively easy to 
collect images exemplifying some visual property, and where on the one
hand we could find images corresponding to a stereotype, but on the
other hand also control images \emph{not} corresponding to the stereotype.

These datasets also contain subsets we call ``neutral''
where stereotypes are not present.
The purpose of these images is to investigate whether LMs are more
sensitive to the choice of images 
in the cases when they cannot rely on stereotypes.

\subsection{The Memory Colors Dataset}

The first dataset is an extension of the \emph{Memory Colors}
dataset \cite{norlund2021transferring}, originally developed for the
purpose of measuring the transfer of information between visual and
textual representations.
The original dataset lists a set of 109 common physical objects, where each
object is listed with a ``memory color'': a stereotypical color we
typically associate with the object. For instance, the dataset lists
tomatoes as stereotypically red although tomatoes frequently have
other colors. The set of objects was annotated by multiple annotators,
and only the objects where there was a perfect or almost perfect
consensus among annotators were included.

The dataset comes with a set of textual templates that can be
used to generate prompts for LMs.
Since the dataset was originally intended for use in LMs where no image was
available, these text templates were intentionally formulated to
elicit stereotypical responses, e.g. \emph{``The typical color of a tomato
is\ldots''}. In our case, we changed the templates to encourage the
model to focus on the image, e.g. \emph{``The color of this tomato
is\ldots''}.

The Memory Colors dataset also includes a set of prototypical images
exemplifying the stereotypical color. For each of the object types, we
collected an additional image where the color was not the stereotypical
one, e.g. a green tomato. 
All images were collected by carrying out a Google image
search and picking the first result.
The majority of objects with unusual colors includes examples of natural
images (e.g. unripe tomatoes, orange sky); in a few cases, the
color had been artifically modified.

We also extended the Memory Colors dataset with 19 neutral
object types selected so that they were not expected to have a stereotypical
color. This set includes common objects such as cars, houses, etc.
We refer to the combined set, including the images with
non-stereotypical colors and the neutral instances, as
the \emph{Extended Memory Colors} dataset.

\subsection{Gender Stereotypes Dataset}
\label{sec:method:gender}

The effect of gender in neural language representation models has been
widely investigated and it is relevant to consider this in multimodal
representations as well.
We compiled a second dataset we term the \emph{Gender Stereotypes}
dataset.
The aim is to identify how good a multimodal model performs in the
prediction of a person's gender when it is fed two different images,
which will act as visual signals for us, one corresponding to a man
and another one corresponding to a woman.
For each pair, there is a sentence that describes the activity.
As in the color dataset, we include stereotypical cases (male-coded and
female-coded, respectively) as well as cases where no stereotype is
present.


The dataset contains 50 different text sentences and 100 images with,
where half of the images show male individuals and half show females.
Internally in the dataset, 19 and 21 text templates were created for
the male and female stereotypical activities, respectively.\footnote{Stereotypical activities were selected from \href{
https://careersmart.org.uk/occupations/equality/which-jobs-do-men-and-women-do-occupational-breakdown-gender}{\color{darkblue}this website}.}
%
Further,
we defined a list of 10 different neutral tasks:
\emph{eating, walking, reading, writing, meditating, talking,
studying, listening to music, clapping, crying}. For these cases, we
assumed that there is no stereotypical gender associated with the
activities.

%
As we will discuss in more detail in Section~\ref{sec:method}, the
property to be predicted will be represented in the sentence as
a \texttt{[MASK]} token to be substituted by a masked LM.
To include an example from the gender stereotype dataset, the sentence
is as \emph{'My therapist is very good,} \texttt{[MASK]} \emph{helped me get
myself together'}; according to the source where we selected the
stereotypical occupations, therapy professionals are more frequently
female.



For each of the 50 text templates, we selected two images, one for
each of the genders. As for the colors dataset, we used the first
result in an image search judged by an annotator to correspond to the
gender in question. We did not take the self-identified gender into
account.

\section{Multimodal Language Models}
\label{sec:models}

The Transformer \cite{vaswani2017attention} is a sequence-based model
that is now the standard architecture in NLP for devising
representation and generation components in neural models.
%
Pre-trained language models such as BERT \cite{devlin2019bert}
based on the architecture of Transformers, have proven capable of
learning powerful representations applicable to a wide range of
tasks. They have yielded state-of-the-art performance in many
downstream tasks.

Multimodal models fusing the textual and visual modalities have been
devised by researchers after looking at the huge success of
pre-trained language models. %
In such models, multiple modalities are considered, and data for the
training of the models is in multiple modalities. As our
research problem revolves around the aspect of multimodality, we will
focus on two modalities: a textual and a visual signal. The visual
signal is in the form of images, and the natural language is the
written text accompanying the images, such as captions or descriptions
of the images.
Examples of such visual/textual Transformers include VilBERT \cite{lu2019vilbert}, LXMERT \cite{tan2019lxmert}, VisualBERT \cite{li2020what}, 
OSCAR \cite{li2020oscar}, ImageBERT \cite{qi2020imagebert}, FLAVA \cite{singh2022flava}, and others.
Most of the earlier models use features extracted from a Faster-RCNN pipeline \cite{ren2015faster}, while later models use visual Transformer architectures \cite{dosovitskiy2021vit}.
These types of models are then trained on datasets that contain text/image pairs such as SBU Captions~\citep{sbu_captions}, MS COCO~\citep{coco}, 
Conceptual Captions~\citep{sharma2018conceptual}, and Visual Genome
QA~\citep{krishnavisualgenome}, using various pre-training tasks. They
are sometimes trained from scratch on the combined language/vision
data and sometimes warm-started from a unimodal model such as BERT. 

For this study, we selected three different multimodal models
to run our experiments on. These image-augmented Transformer models are
\mbox{VisualBERT,} LXMERT, and FLAVA. These three are specifically chosen
to give a certain diversity in the selection of model architecture:
one single-stream CNN-based model, one dual-stream CNN-based model,
and one visual Transformer-based model.

All the models we selected are BERT-like variations that use a 
the technique of Masked Language
Modelling (MLM) during pre-training. This idea was presented in the
original BERT paper \cite{devlin2019bert}. In 
the task of Masked Language Modelling, we predict a token which has
been masked by us in the sentence, given a set of unmasked tokens. In
our case, unmasked tokens are supplemented by the the visual
signals.
The random masking ratio for the MLM is around 15\%, and for
investigation of our experiments one special \texttt{[MASK]} token is
taken.
As we will discuss in Section~\ref{sec:method}, we rely on the ability
of the MLM to predict missing tokens in our experiments.

\paragraph{VisualBERT} This is a single stream multimodal model, i.e,
the language and vision embeddings are processed via a single
Transformer. It is an extension of BERT, by redefining the
process of how input is processed. The language embeddings are
extracted from BERT's tokenizer, which acts as text encoder. 
For the embeddings of the visual signals, Faster-RCNN is used. It
extracts image features in the form of 36 RoI (region of interest)
boxes for each image, and these RoI boxes are used as features.
Each of these 36 ROI boxes are vectors of size 2048. The boxes with
highest probability/confidence are chosen.
The visual representations are
appended at the end of the sequence of word embeddings.

\paragraph{LXMERT}  This model is a dual stream multimodal model,
where the inputs are processed through two Transformers, for natural
language and vision signals respectively. Text is processed in the
same manner as of VisualBERT, based on BERT's tokenizer. The image
features for the LXMERT are extracted by the Faster-RCNN, in the same
way as of VisualBERT, but we also feed the normalized boxes alongside
features, which are locations of these bounding boxes. At last,
the Transformers are fused.

\paragraph{FLAVA}  FLAVA has a text encoder, an image encoder, and a
multimodal encoder. It is a dual stream multimodal model. The text
encoder, has an architecture of ViT (visual Transformers) to extract
single-modal text representations. For the images, an image encoder
based on ViT architecture extracts single-modal image
representations. A separate Transformer, multimodal encoder, is then
applied. The unimodal representations are passed through the fusion
encoder which fuses two modalities, and thus obtaining cross-modal
representations.

\subsection{Model Details}

There is a slight difference in how the two CNN-based models,
VisualBERT and LXMERT, are applied.
In the case of VisualBert, we also input
locations of bounding boxes. For the experiments concerning
VisualBERT, we have used the pretrained BERT
tokenizer,\footnote{bert-base-uncased from the HuggingFace library.} and
VisualBERT with COCO pretraining
checkpoint\footnote{uclanlp/visualbert-vqa-coco-pre from HuggingFace.}
for the model. In the case of LXMERT, the LXMERT base
tokenizer and model\footnote{unc-nlp/lxmert-base-uncased from
HuggingFace library.} were used. For FLAVA, we used the pretrained processor and
model.\footnote{facebook/flava-full from HuggingFace library.}

\section{Methodology of Analysis}
\label{sec:method}

Our benchmarking method uses a cloze-style fill-in-the-blank
approach \cite{petroni2019lmaskb,jiang2020know}, which has previously been applied in experiments investigating the interaction between visual and linguistic representation \cite{norlund2021transferring,hagstrom2022b,hagstrom2022a}.
This approach is easy to apply to BERT-style models that include a
masked language model (MLM) as part of their pre-training pipeline.
When applying the MLM in our experiment, 
the model is provided with an image and a text prompt,
where the visual property to be predicted by the model has been
replaced by the mask dummy token.
We then investigate how well the missing token is predicted under
different circumstances.

Since the nature of the two benchmarks 
is different, we had to apply different methodologies to get the
results. We discuss these details below.

\subsection{The Memory Colors Dataset}

For the Memory Colors dataset, we compare the image having a
stereotypical color to an image with an unusual color for the 
particular object, and to a dummy image containing no meaningful
information.
Following previous work that applied image-augmented LMs to text-only
inputs, we have considered different types of dummy images.
We have used two types of dummy
images: the first one being a completely black image
following \newcite{iki2021effect}, and the second 
consisting of white noise.
However, in experiments we did generally not see major differences
between the behavior of the models when using the black dummy images
and when using the noise images, so we limit the discussion to black
dummy images in the rest of this paper.



For a given text prompt and image, we  mark the output as correctly or
incorrectly predicted depending on whether the
token predicted at the \texttt{[MASK]} position matches 
the color of the label we have provided in the dataset or not.

In these experiments, we did not restrict the output vocabulary to
color terms. In general, after going through the results, it seems
that all the three models tend to output color at the position
of \texttt{[MASK]} token.

\begin{table*}[t!]
\centering
\begin{tabular}{@{}lcccccc}
\toprule
 & \multicolumn{3}{c}{\textbf{Stereotypes}} & & \multicolumn{2}{c}{\textbf{No stereotypes}} \\ \cmidrule(l){2-4}\cmidrule(l){6-7}
\textbf{Model} & \shortstack{\textbf{Original}\\\textbf{image}} & \shortstack{\textbf{Control}\\\textbf{image}} & \shortstack{\textbf{Black}\\\textbf{image}} & & \shortstack{\textbf{Original}\\\textbf{image}} & \shortstack{\textbf{Black}\\\textbf{image}} \\ \midrule
VisualBERT & 0.23 & 0.08 (0.50) & 0.28 (0.41) & & 0.0 & 0.0 (0.84)\\
LXMERT & 0.72 & 0.11 (0.76) & 0.69 (0.87) & & 0.47 & 0.05 (0.47)\\
FLAVA & 0.74 & 0.69 (0.06) & 0.08 (0.08) & & 0.89 & 0.11 (0.11)\\
\bottomrule
\end{tabular}
\caption{Accuracies on the extended Memory Colors datasets. For
control images with unexpected colors, the accuracies are computed with respect to the \emph{new} color, while for the black images the accuracies are with respect to the \emph{original} color. Figures in brackets show the proportion of predictions that are equal to the original prediction.}
\label{tab:colorsresults}
\end{table*}

\subsection{Gender Stereotypes Dataset}

For the Gender Stereotypes dataset, we also consider the output of the
MLM head at the masked position, but in this case we also need to take
into account that 
several words may be applicable in the given context. For this reason,
we create two buckets of male and female words:
\emph{he, male, man, men, boy, his} and \emph{she, female, woman,
women, girl, her}, respectively.
We choose the predicted gender based on the highest probability the elements in the buckets get for the masked token. If the element with the highest probabilty falls in the bucket containing male words, we count this instance as predicted male by the model and vice versa for the female bucket.




\section{Results}
\label{sec:results}

We evaluated the three selected models on the two benchmarks. In both
cases, we compare the predictions when a stereotype is present and the
image corresponds to the stereotype to the case where the
image \emph{does not} correspond to the stereotype.
We also evaluate
cases where there is no stereotype and we carry out similar
comparisons in this case.
Additionally, we look at the model's predictions when provided with a
black dummy image.

\subsection{The Extended Memory Colors Dataset}

Table~\ref{tab:colorsresults} shows the results on the extended Memory
Colors  stereotypes dataset. When using real images, the figures
outside the brackets should be interpreted as predictive accuracies;
for the black dummy images, the figures show the proportions of cases
predicted as the stereotypical color. The figures in brackets show
the proportion of predictions that are identical to the original
prediction.

We note that VisualBERT performs poorly on this dataset, confirming
previously published results that this model is underfitted on visual
data and mostly sticks to the prediction by an equivalent BERT
model.
The effect of the image seems minimal and
its performance is close to the majority-class baseline accuracy
of 0.25. 
%
%

The LXMERT and FLAVA models achive better scores on the original
Memory Colors dataset: both models have accuracies in the 0.70--0.75
range. However, we see clearly that this similarity of performance is
superficial and that the LXMERT model mostly relies on stereotypes:
when we consider the control images with
unexpected colors, the performance of LXMERT is very poor and it
mostly keeps predicting the stereotypical color.
Its performance is somewhat better for the non-stereotypical cases,
but far from perfect.
FLAVA on
the other hand predicts fairly well on the control set, although
somewhat worse than for the images with stereotypical colors; it also
predicts with a good accuracy for the non-stereotypical cases. It is
clear that FLAVA is much more sensitive to the choice of images in
this task.

For the dummy images that are completely black,
the LXMERT model's prediction are again to a large extent identical to the
original predictions.
Again, the FLAVA model is more receptive to the choice of images: it
predicts the color \emph{black} in 92\% of the cases and there is no
discernible effect of stereotypes; it can be discussed whether this is
a desired behavior in this case, since the image does not include an
object of the kind mentioned in the prompt.

Finally, we note that for the non-stereotypical instances, LXMERT's
predictions seem to shift more between the original images and the
black dummy images. This suggests that in cases where the model cannot
rely on a stereotype, the model is more sensitive to the visual input.

\subsection{Gender Stereotypes Dataset}

\begin{table*}[t!]
\centering
\begin{tabular}{lccccccccc}
\toprule
 & \multicolumn{3}{c}{\textbf{Male stereotypes}} &  \multicolumn{3}{c}{\textbf{Female stereotypes}} &  \multicolumn{3}{c}{\textbf{No stereotypes}} \\ \cmidrule(l){2-4}\cmidrule(l){5-7}\cmidrule(l){8-10}
\textbf{Model} & \shortstack{\textbf{Male}\\\textbf{image}} & \shortstack{\textbf{Female}\\\textbf{image}} & \shortstack{\textbf{Black}\\\textbf{image}} &  \shortstack{\textbf{Male}\\\textbf{image}} & \shortstack{\textbf{Female}\\\textbf{image}} & \shortstack{\textbf{Black}\\\textbf{image}} &  \shortstack{\textbf{Male}\\\textbf{image}} & \shortstack{\textbf{Female}\\\textbf{image}} & \shortstack{\textbf{Black}\\\textbf{image}} \\ \midrule
VisualBERT & 0.89 & 0.89 & 0.89 & 0.71 & 0.81 & 0.86 & 0.60 & 0.70 & 0.60\\
LXMERT & 0.84 & 0.68 & 0.73 & 0.95 & 0.76 & 0.90 & 0.90 & 0.40 & 0.80\\
FLAVA & 0.84 & 0.32 & 0.84 & 0.81 & 0.19 & 0.33 & 0.90 & 0.10 & 0.50\\
\bottomrule
\end{tabular}
\caption{Results on the gender stereotypes datasets. The figures
show the proportion predicted as \emph{male}.}
\label{tab:gendersresults}
\end{table*}

Table~\ref{tab:gendersresults} shows the results on the gender
stereotypes dataset. Note that for consistency, the figures
show the proportion of instances predicted
as \emph{male}, so they should not be interpreted as accuracies when
predicting with an image of a female.

Generally speaking, all models tend to predict the \emph{male} class
when provided with an image showing male individuals. When the input
shows a female individual, the picture is more varied.
As in the previous experiment, FLAVA reacts much more strongly to the
choice of images than VisualBERT and LXMERT, and tends to predict
the \emph{male} class for images with males and vice versa.

Unexpectedly, VisualBERT as well as LXMERT both seem to generally
assign higher probabilities to male-coded words, even when the prompt
is stereotypically female; this is surprising since we had expected
these models to predict the stereotypical classes in these cases.
It seems that FLAVA is the only model that shows signs
of \emph{contextual} gender stereotypes 
in this experiment: when provided with a black dummy image, this model
predicts according to what would have been expected stereotypically,
and at 50\% for the non-stereotypical cases.
As we saw in the color experiment, for the non-stereotypical cases
LXMERT seems at least somewhat affected by the choice of images,
although less so than FLAVA.

\section{Related Work}
\label{sec:relwork}

This work falls in the broad category of model
analysis \cite{belinkov2019analysis} of Transformer
models \cite{rogers2020primer}. \newcite{belinkov2019analysis} divide
previous approaches to model analysis into several methodological
categories; in the current work, we use an approach based on
behavioral testing of a specific model behavior.
Specifically, our analysis is based on the outputs of the masked
language model head of BERT-like models, similarly to how 
\newcite{petroni2019lmaskb} and \newcite{jiang2020know} tested BERT models for basic encyclopedic
and commonsense knowledge.

The methodology based on targeted behavioral testing has also been
used to investigate a number of research questions in the analysis of
language-and-vision Transformer models.
In particular, a number of investigations look at what type of
generalizations happen between the visual and textual modalities.
\newcite{cao2020revealing} claimed that when considering attention scores,
the effect of the visual modality is limited and that the textual
modality dominates.
\newcite{norlund2021transferring} investigated the effect of multimodal training on textual representations, and concluded that the degree of transfer between the representations of the respective modalities is limited, at least for CNN-based models;  \newcite{hagstrom2022b,hagstrom2022a} drew similar conclusions based on more extensive experiments that also include the FLAVA model.
%
\newcite{parcalabescu2021seeing} considered the task of predicting
numbers and arrived at a conclusion similar to ours: frequently
occurring numbers are predicted more often by the model.

The previous work that is most closely related to our in terms of
research questions and methodology is that by
\newcite{frank2021vision}. They designed ablation tests where parts of
the image or the text are hidden; as we have discussed, this setup is
comparable to our experiments where black and white-noise images are used.
\newcite{parcalabescu2022valse} introduced the idea of ``foils'': texts
that differs minimally from the one describing the image.
Our use of adversarially selected images can be seen as similar to the idea of foils, but focused on the visual modality.

\section{Conclusions}
\label{sec:conclusions}

In this work, we have proposed a methodological framework based on
controlled tests designed to tease out the influence of stereotypes on
the predictions of visually augmented language models.
The key idea is that we expect common evaluation benchmarks to include
many stereotypical cases that can easily be predicted simply by
relying on language statistics. In order to disentangle the effect of the
stereotype and the contribution of the visual representations
we compare the model's output in cases where the
provided image adheres to the stereotype to cases where it does not.
We also consider the model's behavior in cases where there are no
stereotypes, that is when the prior distribution of outputs is more
evenly distributed.

As an application of this framework, we created two datasets to
facilitate the investigation of stereotypes for two properties: the
color of objects and the gender of people. Each dataset contains a
set of text prompts and corresponding image pairs, where one image in
the pair corresponds to the stereotype and the other is a control
where the stereotypical property is not present. This allows
comparisons to be carried out in a controlled fashion.

Using the two benchmark sets, we evaluated three MLM-based visually
augmented Transformer models: VisualBERT, LXMERT, and FLAVA. There are
clear differences between the models, and in particular some of these
differences emerge much more clearly in the controlled setting.
For instance, the CNN-based LXMERT and Transformer-based FLAVA achieve
similar scores in terms of raw accuracy scores for predicting the
color of objects in images. However, if we consider the control images
where the objects do not have the stereotypical color, the FLAVA
outperforms LXMERT by a wide margin, since LXMERT keeps
predicting the stereotypical color.
%
This means that we can see clear differences among the models with
respect to how sensitive they are to the choice of images.

For the gender stereotypes experiments, the results were somewhat
unexpected since it turned out that the older CNN-based models
almost consistently assigned higher probabilities to male-related
words, where we had expected at least the LXMERT model to be somewhat
affected by stereotypes suggested by the textual prompt. 
The newer FLAVA model on the other hand again predicts more
consistently with the input image in this experiment, and only falls
back on stereotypes when the input images are uninformative.

\subsection{Limitations and Possible Extensions}

As discussed in \S\ref{sec:method:gender}, we have intentionally used a
simplistic operationalization of the notion of gender in this work and selected
images returned by the image search engine when queried for
`male' or `female' respectively, and that the annotator then decided
were prototypical representatives of the male or the female genders.
The self-identified gender of the people in the images was not taken
into account in this experiment and since our goal was to investigate
the sensitivity of visually augmented LMs to the choice of images, it
was a priority to carry out such an evaluation using clear-cut cases.
In a more thorough investigation, it could potentially be useful to
also consider how e.g. the FLAVA model, which seems to be more
affected by the visual input, reacts when presented with images that
do not fall into such clear-cut categories.

The most obvious way that this work could be improved would be to
improve the robustness of the conclusions by scaling up the
investigations along all dimensions: instead of 
considering just the two properties of color and gender, we would like
to investigate a wider selection of properties that would be
meaningful to test in language and vision models. Shape, size, and
orientation are a few possible examples. For each
scenario, it would also be useful to collect more examples than what
we have included here, in order to improve the statistical
robustness. Furthermore, since LMs are sensitive to the choice of a prompt \cite{jiang2020know}, our conclusions would be on firmer ground if we would evaluate on several text prompts for each image.
Naturally, it would be interesting to consider a more extensive selection of models as well.

In this work, we treated the property of being stereotypical as binary
and divided the test cases into groups based on this
property. However, as discussed in the introduction, in reality the
notion of stereotypicality is related to prior probability
distributions.
For this reason, a natural generalization of the experiments we have
carried out here would be to consider stereotypicality on a continuous
scale, e.g. by computing the entropy of the prior distribution and
then to see how this correlates with the probability of incorrect predictions
when encountering an unusual case.

The experiments in this work have been limited to evaluations of the
model's behavior for selected visual-linguistic properties. It remains
to see whether the same idea can be extended beyond evaluation to
devise new \emph{training} methods as well, in order to inject a bias
into the training process aimed at reducing the effects of stereotypes
and encouraging the model to rely on the visual information. This type
of training would typically involve more work in data collection,
unless methods can be devised to adversarially generate images with
unusual properties.

We finally note that the proposed methodology is not limited to the
evaluation of visually augmented LMs, but could be relevant when
considering any extra-linguistic extension of LMs. For instance,
similar pitfalls may occur in the evaluation of LMs augmented with
structural knowledge representations. If a knowledge-augmented LM
correctly predicted some encyclopedic fact \cite{petroni2019lmaskb,jiang2020know}, was
this because of what the knowledge resource contained or because
of text statistics?

\section*{Acknowledgements}

Richard Johansson was supported by the projects \emph{Interpreting and Grounding Pre-trained Representations for NLP} and \emph{Representation Learning for Conversational AI}, both funded by Wallenberg  AI,  Autonomous  Systems  and  Software Program (WASP) funded  by  the  Knut  and  Alice Wallenberg Foundation.

\bibliography{bibliography}

\begin{thebibliography}{27}
\expandafter\ifx\csname natexlab\endcsname\relax\def\natexlab#1{#1}\fi

\bibitem[{Belinkov and Glass(2019)}]{belinkov2019analysis}
Yonatan Belinkov and James Glass. 2019.
\newblock \href {https://doi.org/10.1162/tacl_a_00254} {Analysis methods in
  neural language processing: A survey}.
\newblock \emph{Transactions of the Association for Computational Linguistics},
  7:49--72.

\bibitem[{Bolukbasi et~al.(2016)Bolukbasi, Chang, Zou, Saligrama, and
  Kalai}]{bolukbasi2016}
Tolga Bolukbasi, Kai-Wei Chang, James~Y Zou, Venkatesh Saligrama, and Adam~T
  Kalai. 2016.
\newblock \href
  {https://proceedings.neurips.cc/paper/2016/file/a486cd07e4ac3d270571622f4f316ec5-Paper.pdf}
  {Man is to computer programmer as woman is to homemaker? {D}ebiasing word
  embeddings}.
\newblock In \emph{Advances in Neural Information Processing Systems},
  volume~29. Curran Associates, Inc.

\bibitem[{Cao et~al.(2020)Cao, Gan, Cheng, Yu, Chen, and
  Liu}]{cao2020revealing}
Jize Cao, Zhe Gan, Yu~Cheng, Licheng Yu, Yen-Chun Chen, and Jingjing Liu. 2020.
\newblock \href {https://doi.org/10.1007/978-3-030-58539-6_34} {Behind the
  scene: Revealing the secrets of pre-trained vision-and-language models}.
\newblock In \emph{Computer Vision – ECCV 2020: 16th European Conference,
  Glasgow, UK, August 23–28, 2020, Proceedings, Part VI}, page 565–580,
  Berlin, Heidelberg. Springer-Verlag.

\bibitem[{Devlin et~al.(2019)Devlin, Chang, Lee, and
  Toutanova}]{devlin2019bert}
Jacob Devlin, Ming-Wei Chang, Kenton Lee, and Kristina Toutanova. 2019.
\newblock \href {https://doi.org/10.18653/v1/N19-1423} {{BERT}: Pre-training of
  deep bidirectional transformers for language understanding}.
\newblock In \emph{Proceedings of the 2019 Conference of the North {A}merican
  Chapter of the Association for Computational Linguistics: Human Language
  Technologies, Volume 1 (Long and Short Papers)}, pages 4171--4186,
  Minneapolis, Minnesota. Association for Computational Linguistics.

\bibitem[{Dosovitskiy et~al.(2021)Dosovitskiy, Beyer, Kolesnikov, Weissenborn,
  Zhai, Unterthiner, Dehghani, Minderer, Heigold, Gelly, Uszkoreit, and
  Houlsby}]{dosovitskiy2021vit}
Alexey Dosovitskiy, Lucas Beyer, Alexander Kolesnikov, Dirk Weissenborn,
  Xiaohua Zhai, Thomas Unterthiner, Mostafa Dehghani, Matthias Minderer, Georg
  Heigold, Sylvain Gelly, Jakob Uszkoreit, and Neil Houlsby. 2021.
\newblock \href {https://openreview.net/forum?id=YicbFdNTTy} {An image is worth
  16x16 words: Transformers for image recognition at scale}.
\newblock In \emph{9th International Conference on Learning Representations,
  {ICLR} 2021, Virtual Event, Austria, May 3-7, 2021}. OpenReview.net.

\bibitem[{Frank et~al.(2021)Frank, Bugliarello, and Elliott}]{frank2021vision}
Stella Frank, Emanuele Bugliarello, and Desmond Elliott. 2021.
\newblock \href {https://doi.org/10.18653/v1/2021.emnlp-main.775}
  {Vision-and-language or vision-for-language? {O}n cross-modal influence in
  multimodal transformers}.
\newblock In \emph{Proceedings of the 2021 Conference on Empirical Methods in
  Natural Language Processing}, pages 9847--9857, Online and Punta Cana,
  Dominican Republic. Association for Computational Linguistics.

\bibitem[{Hagstr{\"o}m and Johansson(2022{\natexlab{a}})}]{hagstrom2022b}
Lovisa Hagstr{\"o}m and Richard Johansson. 2022{\natexlab{a}}.
\newblock \href {https://aclanthology.org/2022.coling-1.494} {How to adapt
  pre-trained vision-and-language models to a text-only input?}
\newblock In \emph{Proceedings of the 29th International Conference on
  Computational Linguistics}, pages 5582--5596, Gyeongju, Republic of Korea.
  International Committee on Computational Linguistics.

\bibitem[{Hagstr{\"o}m and Johansson(2022{\natexlab{b}})}]{hagstrom2022a}
Lovisa Hagstr{\"o}m and Richard Johansson. 2022{\natexlab{b}}.
\newblock \href {https://doi.org/10.18653/v1/2022.acl-srw.19} {What do models
  learn from training on more than text? measuring visual commonsense
  knowledge}.
\newblock In \emph{Proceedings of the 60th Annual Meeting of the Association
  for Computational Linguistics: Student Research Workshop}, pages 252--261,
  Dublin, Ireland. Association for Computational Linguistics.

\bibitem[{Iki and Aizawa(2021)}]{iki2021effect}
Taichi Iki and Akiko Aizawa. 2021.
\newblock \href {https://doi.org/10.18653/v1/2021.emnlp-main.167} {Effect of
  visual extensions on natural language understanding in vision-and-language
  models}.
\newblock In \emph{Proceedings of the 2021 Conference on Empirical Methods in
  Natural Language Processing}, pages 2189--2196, Online and Punta Cana,
  Dominican Republic. Association for Computational Linguistics.

\bibitem[{Jiang et~al.(2020)Jiang, Xu, Araki, and Neubig}]{jiang2020know}
Zhengbao Jiang, Frank~F. Xu, Jun Araki, and Graham Neubig. 2020.
\newblock \href {https://doi.org/10.1162/tacl_a_00324} {How can we know what
  language models know?}
\newblock \emph{Transactions of the Association for Computational Linguistics},
  8:423--438.

\bibitem[{Krishna et~al.(2017)Krishna, Zhu, Groth, Johnson, Hata, Kravitz,
  Chen, Kalantidis, Li, Shamma, Bernstein, and Fei-Fei}]{krishnavisualgenome}
Ranjay Krishna, Yuke Zhu, Oliver Groth, Justin Johnson, Kenji Hata, Joshua
  Kravitz, Stephanie Chen, Yannis Kalantidis, Li-Jia Li, David~A. Shamma,
  Michael~S. Bernstein, and Li~Fei-Fei. 2017.
\newblock \href {https://doi.org/10.1007/s11263-016-0981-7} {Visual genome:
  Connecting language and vision using crowdsourced dense image annotations}.
\newblock \emph{Int. J. Comput. Vision}, 123(1):32–73.

\bibitem[{Li et~al.(2020{\natexlab{a}})Li, Yatskar, Yin, Hsieh, and
  Chang}]{li2020what}
Liunian~Harold Li, Mark Yatskar, Da~Yin, Cho-Jui Hsieh, and Kai-Wei Chang.
  2020{\natexlab{a}}.
\newblock \href {https://doi.org/10.18653/v1/2020.acl-main.469} {What does
  {BERT} with vision look at?}
\newblock In \emph{Proceedings of the 58th Annual Meeting of the Association
  for Computational Linguistics}, pages 5265--5275, Online. Association for
  Computational Linguistics.

\bibitem[{Li et~al.(2020{\natexlab{b}})Li, Yin, Li, Zhang, Hu, Zhang, Wang, Hu,
  Dong, Wei et~al.}]{li2020oscar}
Xiujun Li, Xi~Yin, Chunyuan Li, Pengchuan Zhang, Xiaowei Hu, Lei Zhang, Lijuan
  Wang, Houdong Hu, Li~Dong, Furu Wei, et~al. 2020{\natexlab{b}}.
\newblock \href
  {https://www.ecva.net/papers/eccv_2020/papers_ECCV/papers/123750120.pdf}
  {Oscar: Object-semantics aligned pre-training for vision-language tasks}.
\newblock In \emph{European Conference on Computer Vision}, pages 121--137.
  Springer.

\bibitem[{Lin et~al.(2014)Lin, Maire, Belongie, Hays, Perona, Ramanan,
  Doll{\'a}r, and Zitnick}]{coco}
Tsung-Yi Lin, Michael Maire, Serge Belongie, James Hays, Pietro Perona, Deva
  Ramanan, Piotr Doll{\'a}r, and C.~Lawrence Zitnick. 2014.
\newblock \href
  {https://vision.cornell.edu/se3/wp-content/uploads/2014/09/coco_eccv.pdf}
  {Microsoft {COCO}: Common objects in context}.
\newblock In \emph{Computer Vision -- ECCV 2014}, pages 740--755, Cham.
  Springer International Publishing.

\bibitem[{Lu et~al.(2019)Lu, Batra, Parikh, and Lee}]{lu2019vilbert}
Jiasen Lu, Dhruv Batra, Devi Parikh, and Stefan Lee. 2019.
\newblock \href
  {https://proceedings.neurips.cc/paper/2019/file/c74d97b01eae257e44aa9d5bade97baf-Paper.pdf}
  {{ViLBERT}: Pretraining task-agnostic visiolinguistic representations for
  vision-and-language tasks}.
\newblock In \emph{Advances in Neural Information Processing Systems},
  volume~32. Curran Associates, Inc.

\bibitem[{Norlund et~al.(2021)Norlund, Hagstr\"{o}m, and
  Johansson}]{norlund2021transferring}
Tobias Norlund, Lovisa Hagstr\"{o}m, and Richard Johansson. 2021.
\newblock \href {https://aclanthology.org/2021.blackboxnlp-1.10/} {Transferring
  knowledge from vision to language: How to achieve it and how to measure it?}
\newblock In \emph{Proceedings of the Fourth BlackboxNLP Workshop on Analyzing
  and Interpreting Neural Networks for NLP}, pages 149--162, Punta Cana,
  Dominican Republic.

\bibitem[{Ordonez et~al.(2011)Ordonez, Kulkarni, and Berg}]{sbu_captions}
Vicente Ordonez, Girish Kulkarni, and Tamara Berg. 2011.
\newblock \href
  {https://proceedings.neurips.cc/paper/2011/file/5dd9db5e033da9c6fb5ba83c7a7ebea9-Paper.pdf}
  {{Im2Text}: Describing images using 1 million captioned photographs}.
\newblock In \emph{Advances in Neural Information Processing Systems},
  volume~24. Curran Associates, Inc.

\bibitem[{Parcalabescu et~al.(2022)Parcalabescu, Cafagna, Muradjan, Frank,
  Calixto, and Gatt}]{parcalabescu2022valse}
Letitia Parcalabescu, Michele Cafagna, Lilitta Muradjan, Anette Frank, Iacer
  Calixto, and Albert Gatt. 2022.
\newblock \href {https://doi.org/10.18653/v1/2022.acl-long.567} {{VALSE}: A
  task-independent benchmark for vision and language models centered on
  linguistic phenomena}.
\newblock In \emph{Proceedings of the 60th Annual Meeting of the Association
  for Computational Linguistics (Volume 1: Long Papers)}, pages 8253--8280,
  Dublin, Ireland. Association for Computational Linguistics.

\bibitem[{Parcalabescu et~al.(2021)Parcalabescu, Gatt, Frank, and
  Calixto}]{parcalabescu2021seeing}
Letitia Parcalabescu, Albert Gatt, Anette Frank, and Iacer Calixto. 2021.
\newblock \href {https://aclanthology.org/2021.mmsr-1.4} {Seeing past words:
  Testing the cross-modal capabilities of pretrained {V}{\&}{L} models on
  counting tasks}.
\newblock In \emph{Proceedings of the 1st Workshop on Multimodal Semantic
  Representations (MMSR)}, pages 32--44, Groningen, Netherlands (Online).
  Association for Computational Linguistics.

\bibitem[{Petroni et~al.(2019)Petroni, Rockt{\"a}schel, Riedel, Lewis, Bakhtin,
  Wu, and Miller}]{petroni2019lmaskb}
Fabio Petroni, Tim Rockt{\"a}schel, Sebastian Riedel, Patrick Lewis, Anton
  Bakhtin, Yuxiang Wu, and Alexander Miller. 2019.
\newblock \href {https://doi.org/10.18653/v1/D19-1250} {Language models as
  knowledge bases?}
\newblock In \emph{Proceedings of the 2019 Conference on Empirical Methods in
  Natural Language Processing and the 9th International Joint Conference on
  Natural Language Processing (EMNLP-IJCNLP)}, pages 2463--2473, Hong Kong,
  China. Association for Computational Linguistics.

\bibitem[{Qi et~al.(2020)Qi, Su, Song, Cui, Bharti, and
  Sacheti}]{qi2020imagebert}
Di~Qi, Lin Su, Jia Song, Edward Cui, Taroon Bharti, and Arun Sacheti. 2020.
\newblock \href {https://arxiv.org/abs/2001.07966} {Image{BERT}: Cross-modal
  pre-training with large-scale weak-supervised image-text data}.
\newblock \emph{arXiv preprint arXiv:2001.07966}.

\bibitem[{Ren et~al.(2015)Ren, He, Girshick, and Sun}]{ren2015faster}
Shaoqing Ren, Kaiming He, Ross Girshick, and Jian Sun. 2015.
\newblock \href
  {https://proceedings.neurips.cc/paper/2015/file/14bfa6bb14875e45bba028a21ed38046-Paper.pdf}
  {Faster {R-CNN}: Towards real-time object detection with region proposal
  networks}.
\newblock In \emph{Advances in Neural Information Processing Systems},
  volume~28. Curran Associates, Inc.

\bibitem[{Rogers et~al.(2020)Rogers, Kovaleva, and
  Rumshisky}]{rogers2020primer}
Anna Rogers, Olga Kovaleva, and Anna Rumshisky. 2020.
\newblock \href {https://doi.org/10.1162/tacl_a_00349} {A primer in
  {BERT}ology: What we know about how {BERT} works}.
\newblock \emph{Transactions of the Association for Computational Linguistics},
  8:842--866.

\bibitem[{Sharma et~al.(2018)Sharma, Ding, Goodman, and
  Soricut}]{sharma2018conceptual}
Piyush Sharma, Nan Ding, Sebastian Goodman, and Radu Soricut. 2018.
\newblock \href {https://doi.org/10.18653/v1/P18-1238} {Conceptual captions: A
  cleaned, hypernymed, image alt-text dataset for automatic image captioning}.
\newblock In \emph{Proceedings of the 56th Annual Meeting of the Association
  for Computational Linguistics (Volume 1: Long Papers)}, pages 2556--2565,
  Melbourne, Australia. Association for Computational Linguistics.

\bibitem[{Singh et~al.(2022)Singh, Hu, Goswami, Couairon, Galuba, Rohrbach, and
  Kiela}]{singh2022flava}
Amanpreet Singh, Ronghang Hu, Vedanuj Goswami, Guillaume Couairon, Wojciech
  Galuba, Marcus Rohrbach, and Douwe Kiela. 2022.
\newblock \href
  {https://openaccess.thecvf.com/content/CVPR2022/html/Singh_FLAVA_A_Foundational_Language_and_Vision_Alignment_Model_CVPR_2022_paper.html}
  {{FLAVA}: A foundational language and vision alignment model}.
\newblock In \emph{Proceedings of the IEEE/CVF Conference on Computer Vision
  and Pattern Recognition (CVPR)}, pages 15638--15650.

\bibitem[{Tan and Bansal(2019)}]{tan2019lxmert}
Hao Tan and Mohit Bansal. 2019.
\newblock \href {https://doi.org/10.18653/v1/D19-1514} {{LXMERT}: Learning
  cross-modality encoder representations from transformers}.
\newblock In \emph{Proceedings of the 2019 Conference on Empirical Methods in
  Natural Language Processing and the 9th International Joint Conference on
  Natural Language Processing (EMNLP-IJCNLP)}, pages 5100--5111, Hong Kong,
  China. Association for Computational Linguistics.

\bibitem[{Vaswani et~al.(2017)Vaswani, Shazeer, Parmar, Uszkoreit, Jones,
  Gomez, Kaiser, and Polosukhin}]{vaswani2017attention}
Ashish Vaswani, Noam Shazeer, Niki Parmar, Jakob Uszkoreit, Llion Jones,
  Aidan~N Gomez, {\L}ukasz Kaiser, and Illia Polosukhin. 2017.
\newblock \href
  {https://proceedings.neurips.cc/paper/2017/file/3f5ee243547dee91fbd053c1c4a845aa-Paper.pdf}
  {Attention is all you need}.
\newblock In \emph{Advances in Neural Information Processing Systems},
  volume~30. Curran Associates, Inc.

\end{thebibliography}
\end{document}